\documentclass{article}

\PassOptionsToPackage{numbers, compress}{natbib}
\usepackage{rlds_2021}

\usepackage[utf8]{inputenc} %
\usepackage[T1]{fontenc}    %
\usepackage{hyperref}       %
\usepackage{url}            %
\usepackage{booktabs}       %
\usepackage{amsfonts}       %
\usepackage{nicefrac}       %
\usepackage{microtype}      %
\usepackage{listings}
\usepackage{enumitem}
\usepackage{svg}
\usepackage[frozencache,cachedir=.]{minted}
\usepackage{caption}
\usepackage{wrapfig}
\usepackage{subcaption}
\usepackage[bottom]{footmisc}
\newenvironment{code}{\captionsetup{type=listing}}{}
\definecolor{bg}{rgb}{0.95,0.95,0.95}
\definecolor{mygreen}{RGB}{47, 103, 0}
\newcommand{\green}[1]{\textcolor{mygreen}{\texttt{#1}}}

\title{RLDS: an Ecosystem to Generate, Share and Use Datasets in Reinforcement Learning}

\author{%
  Sabela Ramos\thanks{Equal contribution.}\\
  Google Research
  \And
  Sertan Girgin\footnotemark[1]\\
  Google Research
  \And
  L\'eonard Hussenot\\
  Google Research\\
  \And
  Damien Vincent\\
  Google Research
  \And
  Hanna Yakubovich\\
  Google Research
  \And
  Daniel Toyama\\
  Deepmind
  \And
  Anita Gergely\\
  Deepmind
  \And
  Piotr Stanczyk\\
  Google Research
  \And
  Raphael Marinier\\
  Google Research
  \And
  Jeremiah Harmsen\\
  Google Research
  \And
  Olivier Pietquin\\
  Google Research
  \And
  Nikola Momchev\\
  Google Research
}

\begin{document}

\maketitle

\begin{abstract}
We introduce RLDS (Reinforcement Learning Datasets), an ecosystem for recording, replaying, manipulating, annotating and sharing data in the context of Sequential Decision Making (SDM) including Reinforcement Learning (RL), Learning from Demonstrations, Offline RL or Imitation Learning. RLDS enables not only reproducibility of existing research and easy generation of new datasets, but also accelerates novel research. By providing a standard and lossless format of datasets it enables to quickly test new algorithms on a wider range of tasks. The RLDS ecosystem makes it easy to share datasets without any loss of information and to be agnostic to the underlying original format when applying various data processing pipelines to large collections of datasets. Besides, RLDS provides tools for collecting data generated by either synthetic agents or humans, as well as for inspecting and manipulating the collected data. Ultimately, integration with TFDS~\citep{tfds} facilitates the sharing of RL datasets with the research community. 

\end{abstract}

\section{Introduction}\label{sec:intro}

Most Reinforcement Learning (RL) algorithms are notoriously inefficient in the sense that they require large amounts of interactions with their environment to attain optimal performances. To address this issue one trend is to integrate external sources of knowledge. For instance,  Learning from Demonstrations (LfD), Imitation Learning (IL) or Offline RL (ORL) require access to a dataset of logged interactions. Although some datasets are becoming standard~\citep{fu2020d4rl, gulcehre2020rl}, data usage in RL is multifaceted: the variety of tasks at hand is wide. It spans from low data regime (imitation from a single subsampled trajectory \cite{dadashi2020primal}) to high data regime (ORL from large datasets~\citep{agarwal2020optimistic}), from having access to states, actions and rewards~\citep{dadashi2021offline} to having only access to observations
~\citep{torabi2018behavioral}. From expert data~\citep{ho2016generative} to noisy demonstrations~\citep{tangkaratt2020robust}. From low coverage~\citep{dadashi2021offline} to full coverage~\citep{fu2020d4rl} of the state space. From human~\citep{mandlekar2021matters} to synthetic~\citep{gulcehre2020rl} demonstrations. This variety calls for a large diversity of datasets.

With large diversity of datasets and algorithms there emerges the need to compare how different algorithms perform on a given dataset as well as how a given algorithm performs on different datasets. Conducting such experiments requires implementation of (often non-trivial) data pipelines which transform the dataset to the format required by the algorithm (e.g. n-step transitions). Due to the large variety of dataset formats, this requires \textit{ad hoc} implementation for every new type of dataset. Moreover, some experiments are impossible to conduct as the datasets lack the necessary temporal information. E.g., until now, many open-sourced RL datasets~\citep{gulcehre2020rl, ho2016generative, kostrikov2019imitation} are shared as a set of state/action pairs comparable to example/label pairs in supervised learning. Although this format is convenient for some algorithms~\citep{bc}, it discards the temporal information initially present in the data and prevents building methods exploiting such information~\citep{r2d3, smtw, lfl, christiano2017deep, ferret2019self}. Even when temporal information is present in a dataset, what constitutes a step, a transition, and an episode is not always consistent. This may cause errors due to misinterpretation of the data and prevents the community from easily building on top of previous methods and comparing results on existing datasets.

We thus argue that the SDM community would benefit from having a standard way to represent and share datasets, taking into consideration the RL specificities (\textit{e.g.} storing temporal information). This representation should allow each dataset to be used in conjunction with a large variety of algorithms (independent on whether they consume single transitions, n-step sequences, or full episodes) and it should allow practitioners to reuse the same data pipeline for evaluating existing algorithms on new datasets.

In this paper we propose the RLDS ecosystem - a modular suite of tools which aim at addressing the problems described above. More concretely, we describe the following contributions:
\begin{itemize}[leftmargin=*,topsep=1pt,parsep=1pt,partopsep=1pt]
    \item We release \textit{EnvLogger}\footnote{\url{https://github.com/deepmind/envlogger}} and \textit{RLDS Creator}\footnote{\url{https://github.com/google-research/rlds-creator}} which allow for creating synthetic and human generated datasets in a standard format as well as tagging, annotating and inspecting datasets in this format.
    \item We release the \textit{RLDS}\footnote{\url{https://github.com/google-research/rlds}} library of RL-specific transformations which allows for writing efficient and reusable across datasets pipelines to manipulate and transform RLDS datasets and feed them to any SDM algorithm.
    \item We release tools to effortlessly add any dataset created with \textit{EnvLogger} or \textit{RLDS Creator} to Tensorflow Datasets (TFDS)~\citep{tfds}, thus making them easily to discover and access by the community.
    \item We add some of the popular RL datasets ~\citep{gulcehre2020rl, fu2020d4rl} to TFDS in RLDS format, making them easy to consume with RLDS.
    \item We exemplify the full RLDS ecosystem by releasing two new datasets on the Robosuite~\citep{robosuite2020} robotic environment, one gathering synthetic demonstrations recorded with EnvLogger and one with human demonstrations recorded with RLDS Creator.
    \item We release three Notebooks (see Sec.\ref{manipulating}) exemplifying the usage of the RLDS library.
\end{itemize}

\section{Dataset Lifecycle}
We break down the dataset lifecycle in three stages depending on how users interact with the data. RLDS aims at covering each of those stages. We briefly describe them here and in the following sections we discuss the tools we provide to support each stage.

\textbf{Producing the data.} Users provide datasets by recording the interactions with an environment, made by any kind of agent, be it, for example, artificial or human. At this stage, users may want to inspect the data they produce, and either manually or automatically add extra information (e.g. tags) or filter data. Even though users will be able to filter and select the relevant data, we claim that, once selected, this raw data should be stored in a lossless format by recording all the information that is produced, keeping the temporal relation between the data items (e.g., ordering of steps and episodes) without making any assumption on how the dataset is going to be used in the future.
 
\textbf{Consuming the data.}  Researchers use the datasets in order to analyze, visualize or use the data for training machine learning algorithms. Algorithms may consume data in different shapes, not necessarily as it has been stored (for example, some algorithms consume full episodes, others a batch of randomized tuples). To enable this, users need tools to easily set up pipelines transforming raw data into the desired format and allowing to reuse this pipeline across various datasets. Besides, data visualization is an important ingredient for solving machine learning problems that is often neglected in RL. Allowing to visualize the raw data is helpful in most pipelines.

\textbf{Sharing the data.} Datasets are often expensive to produce, and sharing with the wider research community not only enables reproducibility of former experiments, but also accelerates research as it makes it easier to run and validate new algorithms on a range of datasets. Yet, simplifying the sharing of datasets should not affect the ownership  and the credit of the producers. 

\section{Related Work}
\paragraph{Dataset Standardization.}
Gathering datasets in common repositories has long been a concern for the Machine Learning community~\citep{Dua:2019}, yet not always using standard formats. Tools like OpenML \cite{vanschoren2014openml}, or Kaggle provide systems to store datasets and integrate workflows. RLDS is complementary as it enables the creation of systems implementing pipelines reusable across datasets and provides the tools to generate those datasets.

The \mintinline{python}{tf.data} API in Tensorflow~\citep{abadi2016tensorflow} provides a library to manipulate efficiently large streams of data and building modular pipelines. Besides, the Tensorflow Datasets (TFDS)~\citep{tfds} helped standardizing datasets and  establishing good practices for feeding data to machine learning models. HuggingFace Datasets~\citep{huggingface} was built on top of TFDS to provide standard datasets for Natural Language Processing. RLDS builds on top of \mintinline{python}{tf.data} and TFDS and adds explicit support for handling datasets with an episodic structure.

\paragraph{Reinforcement Learning Datasets.}
Within the context of SDM, the usage of external datasets is more recent and no standard format has yet been widely adopted. Recently, with the growing interest of the research community for IL, LfD or ORL, a lot of datasets have been released. They mainly focus on robotics~\citep{fu2020d4rl, toyer2020magical, zeng2020transporter, sharma2018multiple}, some with a particular focus on human data~\citep{mandlekar2021matters, memmesheimer2019simitate, schmeckpeper2020reinforcement}. Some works include datasets for discrete action-environments like games~\citep{gulcehre2020rl, kuttler2020nethack}. Offline Policy Estimation research~\citep{fu2021benchmarks} will likely benefit from such standardization of datasets.

\paragraph{Data Collection Systems.}
MechanicalTurk~\citep{paolacci2010running} is widely adopted for crowdsourcing and data collection, for example, to annotate datasets for supervised machine learning~\citep{squad}. Similarly, RoboTurk~\citep{mandlekar2018roboturk}  provides a tool to crowdsource the collection of data when users manipulate a robotic arm remotely. However, the data is provided with a custom format so users willing to use it in conjunction with data from other tools need a custom pipeline to re-format this data.
\section{Dataset Structure}

An RL dataset is made out of the interactions between an \textit{agent} and an \textit{environment}. Agents can be, for example, RL policies, rule-based controllers, formal planners, humans, animals or a combination of these. More precisely, an RL dataset logically represents a --potentially ordered-- set of \textbf{episodes}, where each episode is a variable-length ordered sequence of interactions (\textbf{steps}) of the agent with the environment. In Section~\ref{sec:intro} we motivated the need of having a standard representation of such datasets. In this section we propose a simple data format which was designed to have the following properties:
\begin{enumerate}[leftmargin=*,noitemsep,topsep=2pt,parsep=2pt,partopsep=2pt]
\item \textbf{lossless}: It preserves all information obtained from interacting with an environment without making any assumptions on how the data is going to be used or transformed. In particular, it maintains the temporal and episodic information. This ensures that each dataset can be used in conjunction with the widest possible variety of algorithms.
\item \textbf{uniform}: It provides a standard way to access the most common fields (e.g. observation, action and reward) returned by RL environments and used in SDM algorithms. It also ensures that those have consistent semantics across all datasets. This provides the ability to implement reusable data pipelines to manipulate and transform those fields.
\item \textbf{flexible}: It allows users to store custom information at the dataset, episode, or step level (for example, weights of the policy used to generate a given episode, or a visual rendering of the environment). Besides, it does not impose any limitations on the shape and type of the data fields (e.g. observations can include ground-truth features, images or nested dictionaries containing both representations). This ensures that the format can be used with any environment and algorithm, including non-standard ones.
\end{enumerate}

RLDS stores sequences of steps where each step contains several fields. To select the set of fields, we looked at classical sources~\citep{suttonbarto}, common environment suites~\citep{brockman2016openai, dmenv, robosuite2020}, existing datasets~\citep{fu2020d4rl, gulcehre2020rl} and common RL frameworks~\citep{acme, TFAgents}. The fields that are included for each step are:
\begin{itemize}[leftmargin=*,noitemsep,topsep=0pt,parsep=0pt,partopsep=0pt]
    \item \green{observation}: the current observation,
    \item \green{action}: action applied to the current observation,
    \item \green{reward}: reward obtained as a result of applying \green{action}.
    \item \green{discount}: discount associated with \green{reward}.
\end{itemize}
Each step and episode may contain custom metadata in the form of additional fields. These fields can be used to store environment-related or model-related metadata, \textit{e.g.} hyperparameters. Besides, each step includes a set of flags standing for step properties:
\begin{itemize}[leftmargin=*,noitemsep,topsep=0pt,parsep=0pt,partopsep=0pt]
    \item \green{is\_first}: indicates if this step is the first of the episode.
    \item \green{is\_last}: indicates that if step is the last of the episode.
    \item \green{is\_terminal}: indicates if the environment considered this step terminal. A time limit wrapper may end a trajectory (\mintinline{python}{is_last=True}) before reaching a terminal state (\mintinline{python}{is_terminal=False}).
\end{itemize}
In a step where \mintinline{python}{is_last=True}, only the observation (and its associated metadata) is defined.

\begin{figure}[h]
\begin{center}
\centerline{\includegraphics[width=0.8\textwidth]{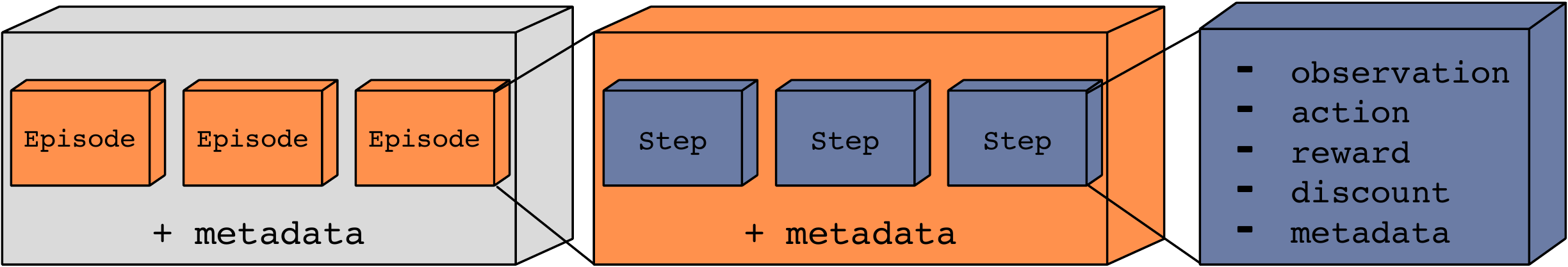}}
\caption{ 
RLDS takes advantage of the inherently standard structure of RL datasets and represents them as a dataset of episodes where each of the episodes contains a nested dataset of steps.}
\label{fig:dataset}
\end{center}
\end{figure}
RLDS stores the data in what is called \textbf{SAR} alignment (to reflect the order State-Action-Reward). However, this is not the only alignment used in the literature and we believe that it is important that we provide equal support for all commonly used step alignments. That is why RLDS provides tools to shift these fields and customize the alignment of the step fields to produce, for example, \textbf{RSA} steps that contain \green{reward}, \green{next\_observation}, \green{next\_action}. 
Note that in this case, the assumptions of what data is defined in each step is different. If we consider the RSA alignment, the first step (where \mintinline{python}{is_first=True}) contains an undefined \green{reward} (and undefined \green{discount}), whereas the last step (\mintinline{python}{is_last=True}) contains defined \green{observation}, \green{reward} and \green{discount}. Similarly, one can easily transform RLDS datasets to the input required by their algorithms, like transitions containing \green{(observation, action, reward, next\_observation)}.

\section{Generating Datasets}
We implement a workflow to record interactions of artificial agents or humans with an environment. Interactions are stored in a lossless format that maintains the temporal relation between steps and episodes. To record synthetic datasets (generated by artificial agents) we release \textit{EnvLogger}, a tool to log agent-environment interactions in an open format. For human-generated datasets, we developed \textit{RLDS Creator}, a tool that enables human demonstrators to interact with environments via a Web UI.

\subsection{Recording datasets with \textit{EnvLogger}}
\setlength\intextsep{0pt}
\begin{wrapfigure}{r}{0.55\textwidth}
\centerline{\includegraphics[width=0.55\textwidth]{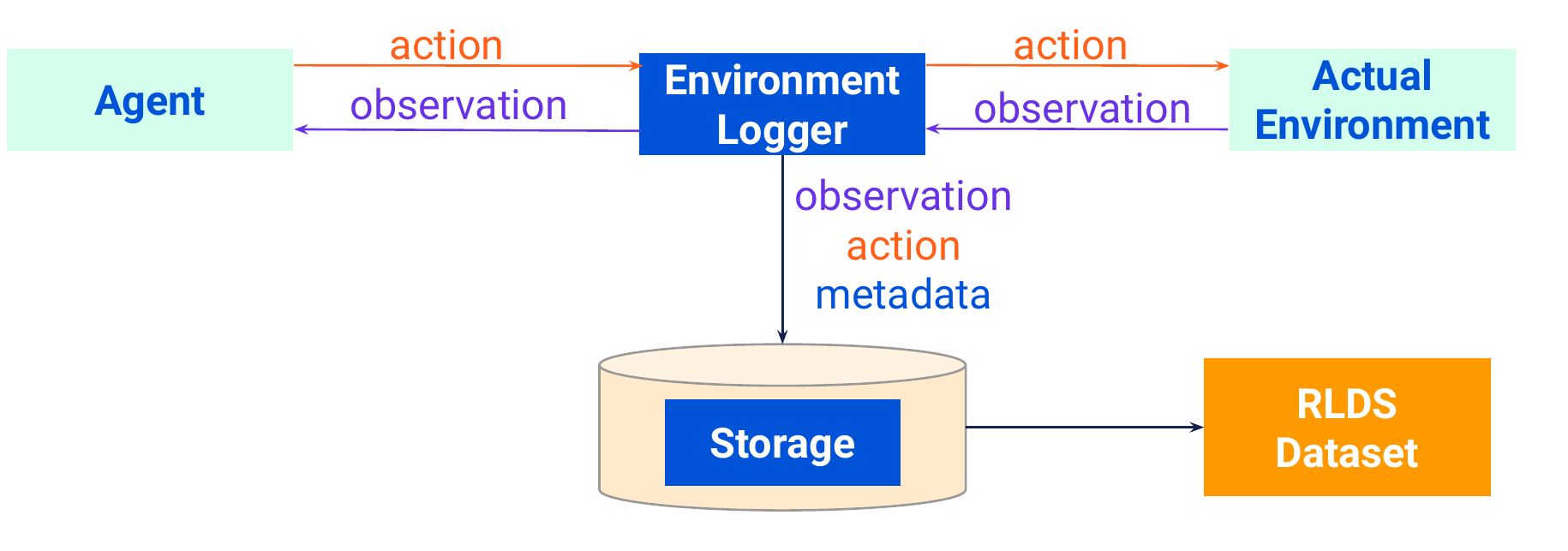}}
\caption{The Environment Logger}
\label{fig:envlogger}
\end{wrapfigure}
EnvLogger is a software library that records agent-environment interactions in an open format and stores these trajectories in long-term storage such as disks. Naturally, it also allows reading them back as entire trajectories, as well as by specific episodes or by individual timesteps. 

EnvLogger also supports storing metadata at the step, episode and dataset level. These may contain any arbitrary data such as neural network logits, gradients, environment statistics and other auxiliary data that can be useful for learning systems and for visualization.

Trajectories are recorded with EnvLogger by wrapping the original environment\footnote{The only requirement is that the environment implements the DM Env~\citep{dmenv} API or is wrapped to support it.}:

\begin{code}
\begin{minted}[
frame=lines,
framesep=2mm,
baselinestretch=1.2,
bgcolor=white,
fontsize=\footnotesize,
linenos
]{python}
env = MyEnvironment(...)
with envlogger.EnvLogger(env, data_directory='/tmp/data') as env:
  env.reset()
  for step in range(1000000):
    action = ...
    timestep = env.step(action)
\end{minted}
\end{code}
EnvLogger is seamlessly integrated in the RLDS ecosystem but, for greater modularity, we designed it to be usable as a stand-alone library as well. Please check App.~\ref{app:envlogger} for more technical details.

\subsection{Collecting and Inspecting Human Datasets: \textit{RLDS Creator}}
\begin{figure}[htb!]
\centering
\includegraphics[width=.9\linewidth]{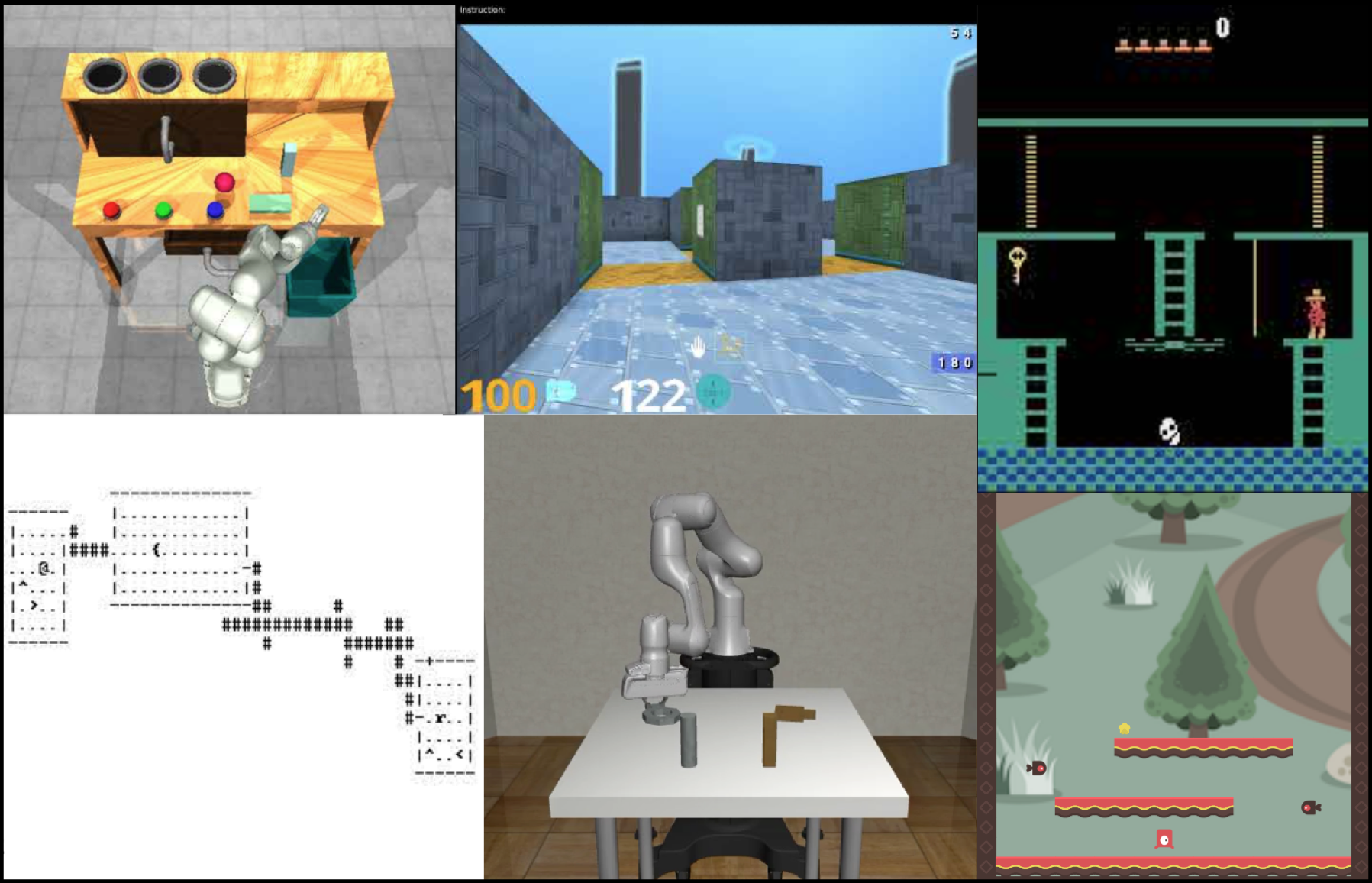}

\caption{Some Environments supported by RLDS Creator: Atari games, DMLab (3D learning environment based on id Software's Quake III Arena), NetHack (single-player text-only roguelike game), Procgen (procedurally-generated 2D games), Robodesk and Robosuite (robot arm).}
\label{fig:creator-envs}
\end{figure}
As in most machine learning settings, collecting human data (e.g. sample trajectories) for RL is a time consuming and labor intensive process. The common approach to address this problem is to use crowd-sourcing, which requires providing a user-friendly access to environments and to scale that access to large numbers of participants. Within the RLDS ecosystem, we release a web-based tool called \textbf{\textit{RLDS Creator}} that provides a universal interface to any human-controllable environment through a browser. RLDS Creator is based on the concept of \textbf{\textit{study}}, a description of the data collection process in which with an online editor the designer specifies:
\begin{itemize}[leftmargin=*,noitemsep,topsep=0pt,parsep=0pt,partopsep=0pt]
    \item the environments and their configuration, e.g. a set of Atari games and their termination conditions,
    \item the instructions for the users, \textit{i.e.} what is the goal of the study and what to do in each environment, \textit{e.g.} to reach a target level or score or to have a specific behavior.
\end{itemize}
The users then can enroll to the study and interact with the environments, e.g. play the Atari games (see Fig.\ref{fig:creator-envs} for other examples) online with their web browser. It is possible to use the keyboard or any of the controllers supported by the standard Gamepad API, including devices such as SpaceMouse\footnote{Some RL environments, such as hand manipulation tasks in Adroit, require specialized input devices (e.g. gloves with sensors) for human interaction; currently, such environments are not supported, but devices with existing browser plugins are in our road-map.}. The interactions, including the user actions, the resulting transitions in the environment, the visual output, \textit{i.e.} rendered images, and other metadata are recorded and stored such that they can be loaded back later using RLDS for analysis or to train agents. RLDS Creator aims at covering a broad set of tasks, spanning from robotics to platform games or text-based games. It focuses on the end-to-end user experience, starting from the study creation to replaying and debugging the collected trajectories, and downloading the resulting datasets with ease. Technical details can be found in App.~\ref{app:creator}.
\section{Sharing}\label{sec:tfds}
One of the goals of standardizing the format of RL datasets is to  share datasets with the research community, increasing reproducibility and enabling running experiments on a wider set of data.
In order to ease data sharing, RLDS is integrated with Tensorflow Datasets (TFDS)~\citep{tfds} - an existing library for sharing datasets within the Machine Learning community. The benefits of using TFDS are:

\textbf{Data Ownership and Access Control:} TFDS does \textit{}{not} host the datasets, it rather enables users to easily download them from the original location. This allows authors to maintain full control over their data. For dataset authors to be fairly credited, all datasets include a citation in their description.

\textbf{Discoverability:} Once the dataset is part of TFDS, it is indexed in the global TFDS catalog, making it accessible to any researcher by using \mintinline{python}{tfds.load(name_of_dataset)}. If users don't want to make their dataset public, they can still benefit from this integration by implementing a TFDS dataset and keeping it private in their own repository.

\textbf{Efficiency and Flexibility:} TFDS is optimized for loading datasets and provides many standard configuration options (for example, it allows to split the episodes into test and training data). Datasets can be read both in Tensorflow and Numpy formats.

\textbf{Support for pre-existing datasets:} TFDS is independent of the underlying format of the original dataset. This means that even if the RL dataset was not generated with our tools, it can still be added to TFDS and used with RLDS (as long as the resulting dataset has an RLDS-compatible format).

In order to integrate RLDS datasets in TFDS, we added support for nested datasets in TFDS, so users are able to load a nested episodic structure. At the moment, TFDS only supports datasets where each of the episodes fit in memory but we intend to remove this limitation.

Users that want to add a dataset to TFDS need to provide a python class that reads from the raw data and produces an RLDS dataset. If the raw data uses the Envlogger format, users can benefit from the RLDS-TFDS integration that provides the base to implement this python 
script. That way, they only need to specify the list of fields and shapes of the data in their dataset (see a full example in App.~\ref{app:tfds}).

At the moment of submitting this paper, the following datasets (compatible with RLDS) are in TFDS\footnote{\url{https://www.tensorflow.org/datasets/catalog/overview}}:
a subset of \textbf{D4RL}~\citep{fu2020d4rl} with tasks from \textbf{Gym Mujoco}~\citep{brockman2016openai} and \textbf{Adroit}~\citep{Rajeswaran-RSS-18}; the \textbf{RLUnplugged}~\citep{gulcehre2020rl} \textbf{DMLab}~\citep{beattie2016deepmind},  \textbf{Atari}~\citep{bellemare13arcade} and \textbf{RWRL}~\citep{rwrl} datasets~\citep{gulcehre2020rl}; two \textbf{Robosuite}~\citep{robosuite2020} datasets generated with the RLDS tools.

We release  two sample datasets on the PickPlaceCan environment of the Robosuite robotic arm simulator \cite{robosuite2020} to illustrate the capabilities of the RLDS ecosystem, from recording datasets using the EnvLogger or the RLDS Creator interface, up to sharing and processing. Those datasets are not meant to supersede previously released datasets \cite{mandlekar2021matters} but rather demonstrate the different functionalities of RLDS. The first dataset is a set of 200 episodes generated by a stochastic agent trained with SAC and recorded using the EnvLogger library. The second dataset is a set of 50 episodes recorded by a single human operator using the RLDS Creator and a gamepad controller. Both datasets hold fixed length episodes of 400 steps. In each episode, a tag is added when the task is completed, this tag is stored as part of the RLDS custom step metadata. With the transformations described in section \ref{manipulating}, one can easily convert the dataset of fixed length episodes into a dataset of variable length episodes ending at the timestep specified by this tag. We also attach, as part of the step metadata, the rendering of the scene which can be useful for visualization or as part of an evaluation procedure.

Additionally, those two datasets incorporate a unique episode ID and agent ID at the episode metadata level to uniquely identify the episode and the agent who generated the episode. We hope the capabilities offered by RLDS will initiate a trend of releasing structured RL datasets, holding more information than just the action/observations.
\section{Loading, Transforming \& Processing}\label{manipulating}
The RLDS format is meant to provide a lossless format in which all the properties of the original dataset are preserved. However, algorithms rarely consume nested datasets of episodes and steps but rather a transformed version like streams of transitions or shuffled n-step transitions.
RLDS and TFDS retrieve the data as \mintinline{python}{tf.data.Dataset} so data can be directly manipulated with \mintinline{python}{numpy} or \mintinline{python}{tf.data} pipelines. However, RLDS provides a library of transformations to ease the handling of the nested datasets of steps. It is optimized for RL scenarios and operations. Using those well tested and optimized transformations, RLDS users (including researchers) have full flexibility to prototype easily some high level functionalities.
Examples of these transformations are:
\begin{itemize}[topsep=0pt,parsep=0pt,partopsep=0pt]
    \item Statistics across the full dataset for selected step fields (or sub-fields).
    \item Flexible batching respecting episode boundaries.
    \item Tools to apply custom transformations to all of the steps of the dataset.
    \item Padding functions and \mintinline{python}{tf.zeros_like} equivalents to construct datasets with empty steps.
    \item Conditional truncation that enables to cut an episode after a step fulfilling a given condition.
    \item Utils to transform the alignment of the steps.
\end{itemize}
Note that RLDS does not provide too high level functions to offer full flexibility on, for example, how to define the train split, but offers the primitives to quickly prototype the desired functionalities.

We illustrate the library of transformations through different scenarios we envision RLDS could be useful for\footnote{See the real examples in \url{https://github.com/google-research/rlds/blob/main/rlds/examples/rlds_examples.ipynb}}. This list is not exhaustive but aims at to showing some of the transformations exposed in the RLDS transformation library. An additional set of examples is released as part of a Notebook\footnote{See tutorial in \url{https://github.com/google-research/rlds/blob/main/rlds/examples/rlds_tutorial.ipynb}}, and performance best practices are available as a separate Notebook\footnote{See performance best practices in \url{https://github.com/google-research/rlds/blob/main/rlds/examples/rlds_performance.ipynb}}

Besides, there are two agents already released~\cite{acme} that make use of RLDS datasets. One of them is the Behavioral Cloning agent~\cite{bc}\footnote{\url{https://github.com/deepmind/acme/blob/master/acme/agents/jax/bc/learning.py}}, that can be fed with a demonstrations iterator coming either from an RLDS dataset or an environment. The other is ValueDICE~\cite{kostrikov2019imitation}\footnote{\url{https://github.com/deepmind/acme/blob/master/examples/gym/run_value_dice.py}}, that uses both an RLDS dataset and online experience from an environment.

\subsection{Example: LfD \& ORL}
We consider the setup where an agent needs to solve a task specified by a reward. We assume a dataset of episodes with the corresponding rewards is available for training. This includes:
\begin{itemize}[leftmargin=*,topsep=0pt,parsep=0pt,partopsep=0pt,noitemsep]
\item The ORL setup~\citep{levine2020offline, wu2019behavior, dadashi2021offline} where the agent is trained solely from the dataset.
\item the LfD setup~\citep{dqfd, ddpgfd, smtw} where, in addition to the dataset, the agent interacts with the environment.
\end{itemize}

Using one of the two provided datasets on the Robosuite PickPlaceCan environment, a typical RLDS pipeline would include the following steps:

\textbf{a.} Deterministically sample $K$ episodes from the dataset so the performance of the agent is expressed as a function of the number of available episodes.
\begin{code}
\begin{minted}[
frame=lines,
framesep=2mm,
baselinestretch=1.2,
bgcolor=white,
fontsize=\footnotesize,
linenos
]{python}
ds = tfds.load('robosuite_panda_pick_place_can')
ds = ds.shuffle(30, seed=42, reshuffle_each_iteration=False)
ds = ds.take(K)
\end{minted}
\end{code}
\textbf{b.} Combine the observations used as an input of the agent. The Robosuite datasets include many fields in the observations and one could try to train the agent from the state or from the visual observations.
\begin{code}
\begin{minted}[
frame=lines,
framesep=2mm,
baselinestretch=1.2,
bgcolor=white,
fontsize=\footnotesize,
linenos
]{python}
def prepare_observation(step):
  """Filters the obseravtion to only keep the state and flattens it."""
  observation_names = ['robot0_proprio-state', 'object-state']
  step[rlds.OBSERVATION] = tf.concat(
      [step[rlds.OBSERVATION][key] for key in observation_names], axis=-1)
  return step
 
dataset = rlds.transformations.map_nested_steps(dataset, prepare_observation)
\end{minted}
\end{code}
 \textbf{c.} Finally, convert the dataset of episodes into a dataset of transitions that can be consumed by algorithms such as SAC~\citep{haarnoja2018soft} or TD3~\citep{fujimoto2018addressing}.

\begin{code}
\begin{minted}[
frame=lines,
framesep=2mm,
baselinestretch=1.2,
bgcolor=white,
fontsize=\footnotesize,
linenos
]{python}
def batch_to_transition(batch):
  """Converts a pair of consecutive steps to a custom transition format."""
  return {'s_cur': batch[rlds.OBSERVATION][0],
          'a': batch[rlds.ACTION][0],
          'r': batch[rlds.REWARD][0],
          's_next': batch[rlds.OBSERVATION][1]}

def make_transition_dataset(episode):
  """Converts an episode of steps to a dataset of custom transitions."""
  # Create a dataset of 2-step sequences with overlap of 1.
  batched_steps = rlds.transformations.batch(episode[rlds.STEPS], size=2, shift=1)
  return batched_steps.map(batch_to_transition)

transitions_ds = dataset.flat_map(make_transition_dataset)
\end{minted}
\end{code}
This example highlights the synergy of the RLDS transformations with existing functions of the \mintinline{python}{tf.data.Dataset} library. Creating the test split only relies on standard \mintinline{python}{tf.data.Dataset} functions while a convenience function \mintinline{python}{map_nested_steps} to work on nested datasets is used to create the desired observation.

\subsection{Example: Absorbing Terminal States in IL}

Imitation learning is the setup where an agent is trained to mimic a behavior from the observation of some sample episodes of that behavior.
In particular, the reward is not specified.

The data processing pipeline involves every piece seen in the LfD setup (create a train split, assemble the observation, ...) but also has some specifics.
One specific is related to the particular role of the terminal state in imitation learning.
In standard RL tasks, looping over the terminal states only brings zero in terms of reward. In imitation learning, making this assumption of zero reward for transitions from a terminal state to the same terminal state induces some bias in algorithms like GAIL. %
One way to counter this bias was proposed in \cite{kostrikov2018discriminator}. It consists in learning the reward value of the transition from the terminal state to itself.
Implementation wise, to tell a terminal state from another state, an `absorbing' bit is added to the observation (1 for a terminal state, 0 for a regular state). The dataset is also augmented with transitions from terminal state to itself so the agent can learn from them. In the code below we chose to set the action and observation of those transitions to zeros but a reasonable alternative would be to keep the observation unchanged or to generate random actions.
\begin{code}
\begin{minted}[
frame=lines,
framesep=2mm,
baselinestretch=1.2,
bgcolor=white,
fontsize=\footnotesize,
linenos
]{python}
def duplicate_terminal_step(episode):
  """Duplicates the terminal step if the episode ends in one. Noop otherwise."""
  return rlds.transformations.concat_if_terminal(
      episode, make_extra_steps=tf.data.Dataset.from_tensors)

def convert_to_absorbing_state(step):
  padding = step[rlds.IS_TERMINAL]
  if step[rlds.IS_TERMINAL]:
    step[rlds.OBSERVATION] = tf.zeros_like(step[rlds.OBSERVATION])
    step[rlds.ACTION] = tf.zeros_like(step[rlds.ACTION])
    # This is no longer a terminal state as the episode loops indefinitely.
    step[rlds.IS_TERMINAL] = False
    step[rlds.IS_LAST] = False
  # Add the absorbing bit to the observation.
  step[rlds.OBSERVATION] = tf.concat([step[rlds.OBSERVATION], [padding]], 0)
  return step

absorbing_state_ds = rlds.transformations.apply_nested_steps(
    dataset, duplicate_terminal_step)
absorbing_state_ds = rlds.transformations.map_nested_steps(
    absorbing_state_ds, convert_to_absorbing_state)
\end{minted}
\end{code}
\subsection{Example: Data Analysis}
One significant use case we envision for RLDS is the analysis of collected datasets.
There is no standard data analysis procedure as what is possible is only limited by the imagination of the users. We expose in this section a fictitious use case to illustrate how custom tags stored in a RL dataset can be processed as part of an RLDS pipeline.
Let's assume we want to compute the returns of the episodes present in the dataset of human episodes on the robosuite PickPlaceCan environment. This dataset holds episodes of fixed length of size 400 but also has a tag to indicate the actual end of the task, i.e. when the can is placed in a bin.
Let's consider that we want the returns of the variable length episodes ending on the completion tag.

\begin{minted}[
frame=lines,
framesep=2mm,
baselinestretch=1.2,
bgcolor=white,
fontsize=\footnotesize,
linenos
]{python}
def placed_tag_is_set(step):
  return tf.not_equal(tf.math.count_nonzero(step['tag:placed']),0)

def compute_return(steps):
  """Computes the return of the episode up to the 'placed' tag."""
  # Truncate the episode after the placed tag, then sum 
  steps = rlds.transformations.truncate_after_condition(
      steps, truncate_condition=placed_tag_is_set)
  return rlds.transformations.sum_dataset(steps,
                                          lambda step: step[rlds.REWARD])

returns_ds = dataset.map(lambda episode: compute_return(episode[rlds.STEPS]))
\end{minted}
Similarly, one might want to compare the reward distribution of several datasets. Notice that, without RLDS, one would have to treat differently each of the datasets coming from Robosuite, RLUnplugged and D4RL, while they go through the exact same processing pipeline here:
\begin{minted}[
frame=lines,
framesep=2mm,
baselinestretch=1.2,
bgcolor=white,
fontsize=\footnotesize,
linenos
]{python}
import matplotlib.pyplot as plt

environments_names = [['d4rl_...', 'd4rl_...', 'd4rl_...'],
                       ['rlu_...', 'rlu_...'],
                       ['robosuite_...', 'robosuite_...']]
fig, axs = plt.subplots(1, len(environments_groups))

for i, group in enumerate(environments_names):
  for dataset_name in group:
    # Load 10 episodes of each dataset
    ds = tfds.load(dataset_name, split = 'train[:10]')
    # Flatten episodes and keep only reward
    ds = ds.flat_map(lambda episode: episode[rlds.STEPS])
    ds = ds.map(lambda x: x[rlds.REWARD])
    axs[i].hist(list(ds.as_numpy_iterator()), label=dataset_name)
    axs[i].legend()

plt.plot()
\end{minted}

\begin{figure}[bht]
    \centering
    \includegraphics[width=\linewidth]{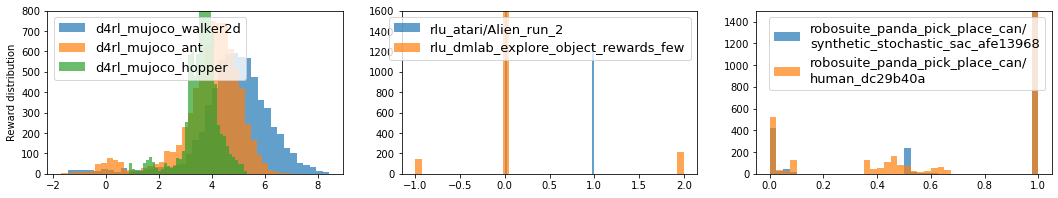}
    \caption{The histograms produced by the code snippet above.}
    \label{fig:hist}
\end{figure}

\section{Conclusion}

We release RLDS, the first ecosystem for datasets meant to accelerate research in SDM. RLDS proposes a standard format to store datasets without loss of information, especially the temporal dependencies between steps and between episodes, and allows to reuse data processing pipelines across multiple datasets. To ease the process of manipulating and generating data, RLDS includes a library of common RL transformations. Finally, it integrates with TFDS in order to ease the sharing of the data and ensures that authors keep the ownership and credit for their data. Pre-existing datasets (or datasets generated outside of the RLDS tooling), can also benefit from the ecosystem by adding the dataset to TFDS. We claim that the RLDS ecosystem not only improves reproducibility, but also enables new research by making it easier to share and reuse data in SDM problems.

\section*{Acknowledgements}
The authors want to thank the collaboration of other engineers and researchers that provided feedback and contributed to the project. In particular, George Tucker, Sergio Gomez, Jerry Li, Caglar Gulcehre, Pierre Ruyssen, Etienne Pot, Anton Raichuk, Gabriel Dulac-Arnold, Nino Vieillard, Matthieu Geist, Alexandra Faust, Eugene Brevdo, Tom Granger, Zhitao Gong and Tom Small.

\bibliographystyle{abbrvnat}
\bibliography{main}

\newpage
\appendix
\section{New Assets}\label{app:assets}

With this paper, we are introducing two new datasets generated with the robosuite robotic arm simulator \citep{robosuite2020}. In Section~\ref{sec:tfds} we document its content as well as the generation details. Detailed documentation is available in the TFDS catalog\footnote{\url{https://www.tensorflow.org/datasets/catalog/robosuite_panda_pick_place_can}}. It can be downloaded and inspected following the instructions in the TFDS website or using our python Notebooks \footnote{\url{https://github.com/google-research/rlds/blob/main/rlds/examples/rlds_tutorial.ipynb}} \footnote{\url{https://github.com/google-research/rlds/blob/main/rlds/examples/rlds_examples.ipynb}} \footnote{\url{https://github.com/google-research/rlds/blob/main/rlds/examples/_performance.ipynb}}.

Since TFDS doesn't host the data, we made it publicly available in Google Cloud Storage \footnote{\url{https://console.cloud.google.com/storage/browser/rlds_external_data_release}} with a Creative Commons (CC-BY) license\footnote{\url{https://storage.googleapis.com/rlds_external_data_release/README.md}}. The authors bear all responsibility in case of violation of rights.

The other datasets mentioned in Sec.~\ref{sec:tfds} (D4RL \citep{fu2020d4rl}, RL Unplugged DMLab~\citep{gulcehre2020rl}, RL Unplugged Atari \citep{gulcehre2020rl} and RL Unplugged RWRL~\citep{rwrl}) have been added by us to the TFDS catalog\footnote{\url{https://www.tensorflow.org/datasets/catalog/overview}} in agreement with the original authors. Besides, they keep the ownership and attribution, and the data maintains its original license.

None of the datasets discussed include any personally identifiable information or offensive content.

Regarding the sofware, the three new packages are available on GitHub and distributed with an Apache 2.0 license:

\begin{itemize}
    \item RLDS: \url{https://github.com/google-research/rlds}
    \item RLDS Creator: \url{https://github.com/google-research/rlds-creator}
    \item EnvLogger: \url{https://github.com/deepmind/envlogger}
\end{itemize}

The repositories include documentation and instructions on how to use them.

As part of this work, we made contributions to the TFDS library, that is also available on GitHub (\url{https://https://github.com/tensorflow/datasets}) with an Apache 2.0 license.

The authors are committed to maintain and support the software released in GitHub and the published datasets.

\section{EnvLogger - details}\label{app:envlogger}
EnvLogger being a stand-alone library allows for building custom data processing tools which directly operate on the output of EnvLogger. For this reason, the data is stored in industry standard protocol buffers in an open source file format called Riegeli~\citep{riegeli}. This choice was driven by:
\begin{enumerate}[topsep=0pt,parsep=0pt,partopsep=0pt]
    \item Interoperability of programming languages: protocol buffers can be read and produced by a number of popular programming languages and tools. Data can be produced in one language and consumed in another.
    \item Future-proof and open: Since the format is completely open sourced, consumption of the produced data is not tied to the EnvLogger library. New libraries and programming languages may be used to produce and read data in the same format.
    \item Efficiency: Protocol buffers and Riegeli are binary formats which provide greater performance than text-based formats such as JSON or CSV. Riegeli also has built-in support for compression, further reducing storage requirements.
    \item Good trade-off between fast writes and reads: Riegeli is a natural logging format with very fast appends while it also supports reliable seeks when using an external index.
    \item Server-less: data can be produced and read without the need of a separate server or database.
\end{enumerate}
Please refer to EnvLogger’s code repository\footnote{\url{https://github.com/deepmind/envlogger}} for more technical information. 

\section{RLDS Creator - details}\label{app:creator}

\begin{figure}
    \centering
    \includegraphics[width=0.6\linewidth]{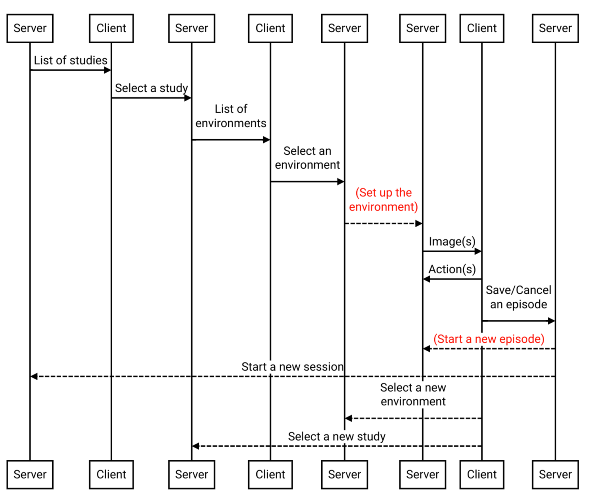}
    \caption{The communication protocol between the client and the server in RLDS Creator.}
    \label{fig:creator-protocol}
\end{figure}

RLDS Creator follows the distributed client-server model, with the (web) servers responsible for the core logic, i.e. managing the studies and simulating the environments based on user actions, and clients displaying the visual representations of the environment to the user and sending user actions, e.g. controller input, to the servers. This makes it possible to support a wide range of environments, including those with conflicting dependencies (by running them under separate processes in the server) or high resource requirements (e.g. hardware rendering). Examples of environments supported by the RLDS Creator are presented in Fig.~\ref{fig:creator-envs}. Clients and servers communicate using asynchronous bi-directional messaging over HTTP (WebChannel or WebSockets); Fig. \ref{fig:creator-protocol}.

\begin{figure}
    \centering
    \includegraphics{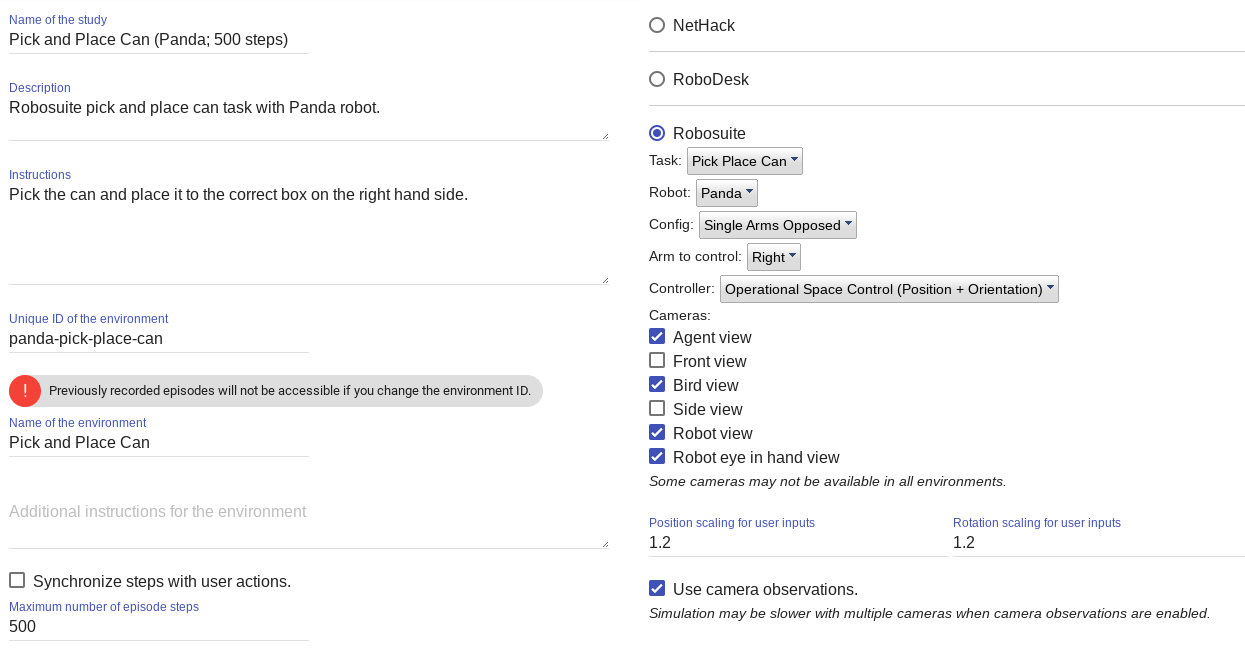}
    \caption{RLDS Creator study editor. A study has environment agnostic, e.g. name and instructions (left), and environment specific settings, e.g. name of the robot and task for Robosuite (right).}
    \label{fig:creator-study-editor}
\end{figure}

\begin{figure}
    \centering
    \includegraphics[width=\linewidth]{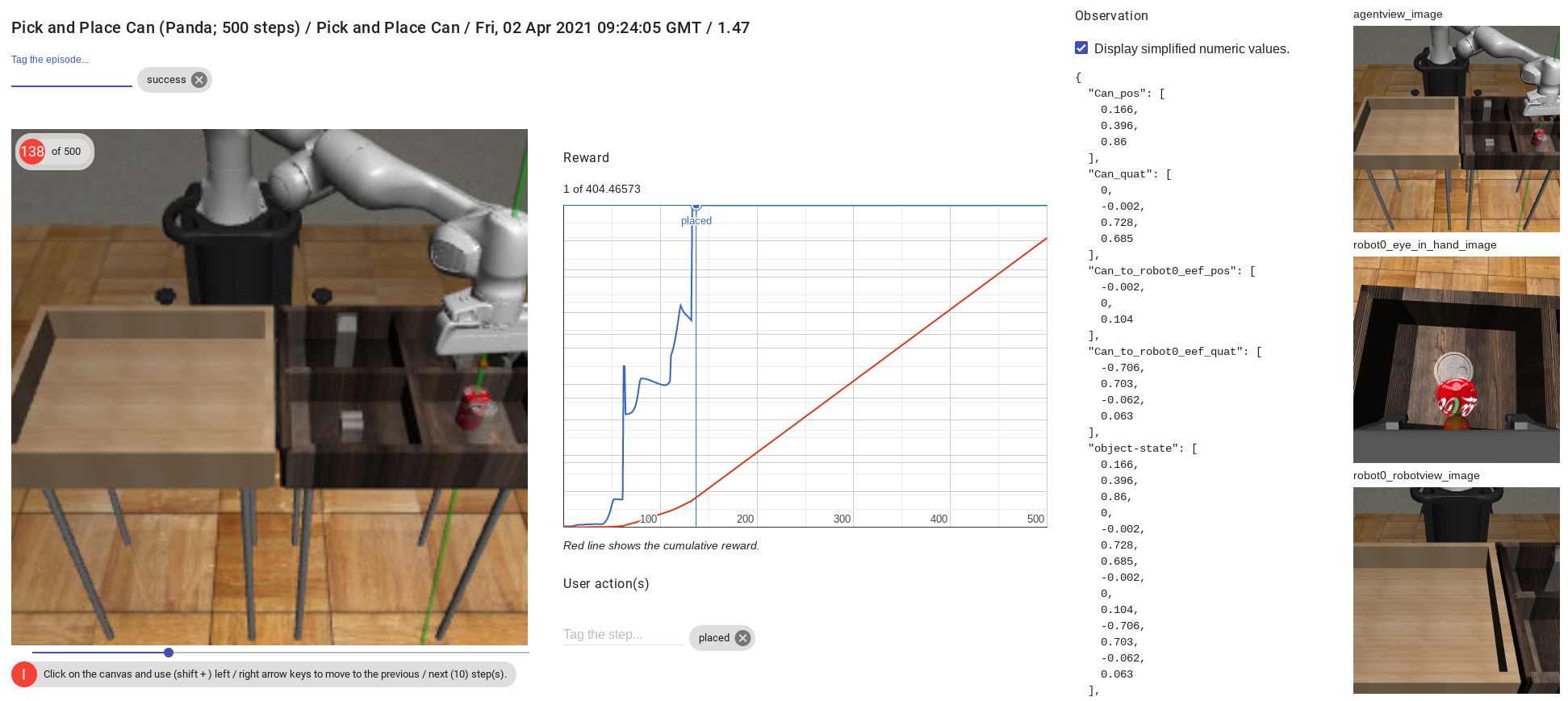}
    \caption{The replay functionality of RLDS Creator allows to examine a recorded episode step by step. Both the visual representation of the environment as seen by the user (left) and the actual environment observations (both numeric and image features are supported; right) are presented. The reward profile (middle) displays a time-series of the step rewards and helps to pinpoint important events (in this example, picking the soda can and placing it to the correct bin result in two sudden jumps in the received rewards). Individual steps (bottom middle) and the episode itself can be tagged (top left); these tags are exported to the RLDS dataset and can be used e.g. for filtering purposes when training a policy.}
    \label{fig:creator-replay}
\end{figure}

In RLDS Creator, \textbf{the regular workflow} for collecting human data is as follows:

\textbf{The study designer}:
\begin{enumerate}
    \item Creates a data collection study using an online editor (Fig. \ref{fig:creator-study-editor}). Each study consists of one or more environments from the same domain (i.e. with compatible observation and action spaces) but possibly with different settings, e.g. "pick and place" task in Robosuite with various objects or a set of platform games in Procgen. The instructions for the users, i.e. what the goal of the study is and what to do in each environment, are specified in free-form text. Both synchronous (environment advances to the next step on user action) and asynchronous (environment advances at a fixed frame rate regardless of user action) modes are supported. 
    \item Tests the study by recording sample trajectories (see below). Using the built-in replay functionality of RLDS Creator, it is possible to visually inspect the (environment) observation, (user) action and reward at each step (Fig. \ref{fig:creator-replay}). This helps to pinpoint possible issues (e.g. reward may be dense instead of sparse or some features may be missing in the observations) and fix them early, thus avoiding throwing away collected data later.
    \item Starts the study; this creates a permalink.
\end{enumerate}

\begin{figure}
    \centering
    \includegraphics{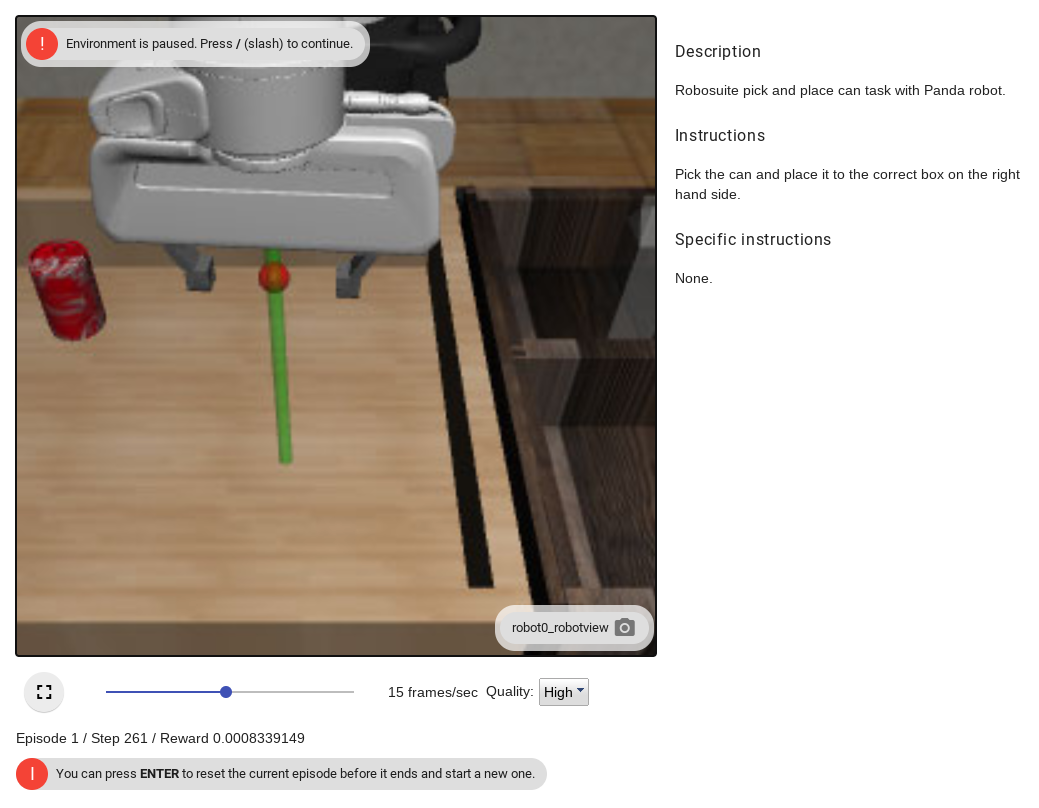}
    \caption{On the environment player of RLCS Creator, the user can read the instructions for the study (right), see the visual representation of the environment (left) and interact with it using different input devices. It is possible to pause the simulation, change the frame rate and image quality, enter the full screen mode for a more immersive experience or switch to different cameras if supported by the environment.}
    \label{fig:creator-player}
\end{figure}

\textbf{A user} opens permalink (enrolling the study) and starts a session to interact with the defined environments. In a session,
\begin{enumerate}
    \item User selects an environment in the study and views the instructions for the study and environment, if any.
    \item Starts a new episode (or trajectory) and interacts with the environment, i.e. sees a visual representation of the current state and takes actions (Fig. \ref{fig:creator-player}).
    \item At any point during the interaction loop, the user can (i) pause the environment temporarily; if it is not unpaused after some time, the session will expire and the episode will be marked as "abandoned", or (ii) cancel the episode; the episode will be marked as "canceled". 
    \item When the episode ends, the user is asked for confirmation to save it. The episode will be marked as "completed" if confirmed, otherwise as "rejected". After this, the user can end the session, continue with a new episode of the current environment, or choose another environment in the study.
\end{enumerate}

\begin{figure}
    \centering
        \begin{subfigure}{0.49\textwidth}
        \includegraphics[width=\linewidth]{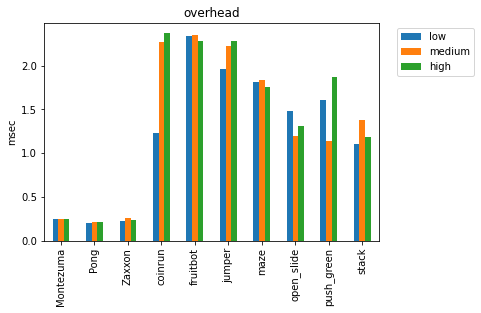}
        \end{subfigure}
        \begin{subfigure}{0.49\textwidth}
        \includegraphics[width=\linewidth]{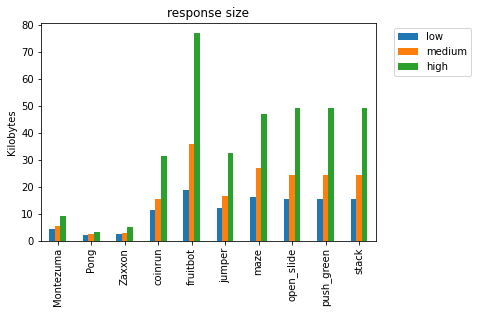}
        \end{subfigure}
    \caption{Image encoding and video recording overhead in a sample set of Atari, Procgen and Robodesk environments varies from between 0.5-2 msecs based on the resolution (left); this is also correlated with the size of the data sent to the client at each step (right).}
    \label{fig:creator-stats}
\end{figure}

On a standard workstation, the overhead of RLDS Creator (i.e. mainly encoding of the images and recording of the videos) is between 0.5-2.0 msecs for each step depending on the complexity of the environment and the resolution of the images; this is also correlated with the size of the data that is sent to the client (Fig. \ref{fig:creator-stats}).

\textbf{Both the study designer and enrolled users} can replay recorded trajectories and provide additional metadata for them. When replaying a trajectory, it is possible to walk through its steps and visually inspect the observations, user and environment actions and the reward at each step. In particular, the reward profile feature (i.e. step rewards as a time series) makes it easy to pinpoint important events in the trajectory (Fig. \ref{fig:creator-replay}). The metadata can be for the trajectory itself (e.g. indicating "success" or "failure" if this is not explicit in the environment) or for a particular step (e.g. when an object was "picked" or "placed"). The "tagging" functionality of the replay tool allows entering such metadata in a structured and accessible form, which can later be used easily in RLDS, e.g. for filtering purposes; free-text notes are also supported (Fig. \ref{fig:creator-replay}). 

\begin{figure}
    \centering
    \includegraphics[width=0.9\linewidth]{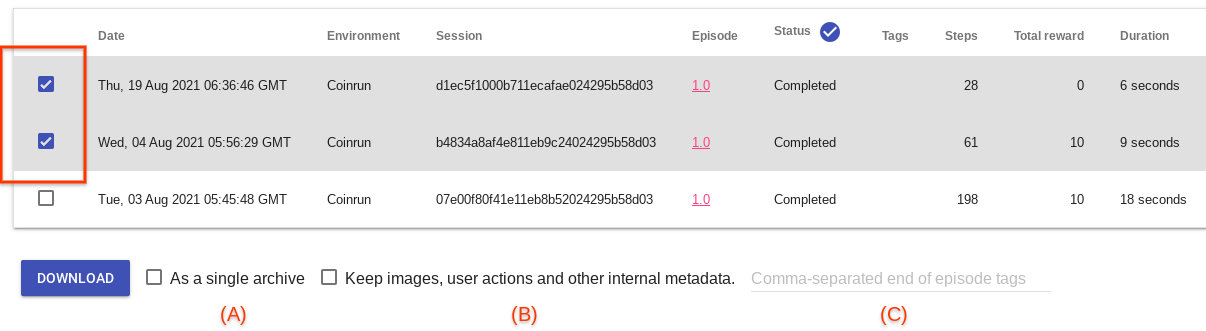}
    \caption{To create a dataset from the recorded episodes, the user can simply query and select them using the RLDS Creator. It is possible to download the dataset as a single archive file (A) strip internal metadata (e.g. rendered images; B) and truncate the episodes based on the tags that were added using the replay tool (C).}
    \label{fig:creator-download}
\end{figure}

Finally, \textbf{the study designer} browses the recorded trajectories for the study, selects (a subset) and downloads them as an RLDS dataset (Fig. \ref{fig:creator-download}).

\newpage
\section{TFDS: adding an RLDS-based dataset}\label{app:tfds}
This section shows an example of a script to add an RLDS dataset (recorded with Envlogger or RLDS Creator) to TFDS. It is based on a real example from TFDS (\url{https://github.com/tensorflow/datasets/blob/master/tensorflow_datasets/rlds/robosuite_panda_pick_place_can/robosuite_panda_pick_place_can.py}).

Note that datasets recorded with a different method can still be added but will need to reimplement some parts of the \green{rlds\_base} module.

\begin{code}
\begin{minted}[
frame=lines,
framesep=2mm,
baselinestretch=1.2,
bgcolor=white,
fontsize=\footnotesize,
linenos
]{python}
import tensorflow.compat.v2 as tf
import tensorflow_datasets.public_api as tfds
from tensorflow_datasets.rlds import rlds_base

_DESCRIPTION = """
Long dataset description
"""
_CITATION = """
BibTeX citation of the original work that generated this dataset
"""

class DatasetName(tfds.core.GeneratorBasedBuilder):

  VERSION = tfds.core.Version('1.0.0')
  RELEASE_NOTES = {
      '1.0.0': 'Initial release.',
  }
  
  _DATA_PATHS = {
      'my_dataset': '<path to download the data from>'
  }

  # The configurations allow to generate different versions of this dataset. Each of the versions
  # need to specify the data types and shapes of the content of the dataset.
  BUILDER_CONFIGS = [
      rlds_base.DatasetConfig(
          name='my_dataset_config_name',
          observation_info={
              'object-state': tfds.features.Tensor(shape=(14,), dtype=tf.float64),
              # ...
          },
          action_info=tfds.features.Tensor(shape=(7,), dtype=tf.float64),
          reward_info=tf.float64,
          discount_info=tf.float64,
          episode_metadata_info={
              'agent_id': tf.string,
              'episode_index': tf.int32,
              'episode_id': tf.string,
          },
          step_metadata_info={
              'tag:placed': tf.bool,
              'image': tfds.features.Image(),
          },
          citation=_CITATION,
          overall_description=_DESCRIPTION,
          description='Human generated dataset.',
          supervised_keys=None,
      ),
  ]

  def _info(self) -> tfds.core.DatasetInfo:
    """Returns the dataset metadata."""
    return rlds_base.build_info(self.builder_config, self)

  def _split_generators(self, dl_manager: tfds.download.DownloadManager):
    """Returns SplitGenerators to comply with the TFDS API."""
    path = dl_manager.download_and_extract({
        'file_path': self._DATA_PATHS[self.builder_config.name],
    })
    return {
        'train': self._generate_examples(path),
    }

  def _generate_examples(self, path):
    """Yields examples. In this case, episodes."""
    file_path = path['file_path']
    return rlds_base.generate_beam_examples(file_path)
\end{minted}
\caption{Script to add an RLDS dataset to TFDS.}
\end{code}

\end{document}